\documentclass[10pt,journal,compsoc]{IEEEtran}

\usepackage{times}
\usepackage{epsfig}
\usepackage{graphicx}
\usepackage{amssymb}
\usepackage[tight]{subfigure}
\usepackage{caption}
\usepackage{subfig}
\usepackage{multirow}
\usepackage{comment}
\usepackage{caption}
\usepackage[shortlabels]{enumitem}
\usepackage{array}
\usepackage{threeparttable}
\usepackage{amsmath}
\usepackage{url}

\makeatletter
\def\footnoterule{\relax%
 \kern-5pt
 \hbox to \columnwidth{\hfill\vrule width \columnwidth height 0.6pt\hfill}
 \kern4.6pt}
\makeatother

\newcolumntype{P}[1]{>{\centering\arraybackslash}p{#1}}
\newcolumntype{M}[1]{>{\centering\arraybackslash}m{#1}}

\usepackage{algorithm}
\usepackage[noend]{algpseudocode}
\makeatletter
\def\BState{\State\hskip-\ALG@thistlm}
\makeatother

\setlist[enumerate]{nosep}
\setlist[itemize]{nosep}

\usepackage{multirow}
\usepackage{hhline}%
\usepackage{booktabs}
\usepackage{tabularx}

\DeclareMathOperator{\IR}{\mathbb{R}}
\DeclareMathOperator{\EL}{\mathcal{L}}
\DeclareMathOperator{\mcF}{\mathcal{F}}

\DeclareMathOperator{\bfi}{\mathbf{f}_i}
\DeclareMathOperator{\bwj}{\mathbf{w}_j}

\DeclareMathOperator{\bwjb}{\mathbf{w}_j^{batch}}

\ifCLASSOPTIONcompsoc
  \usepackage[nocompress]{cite}
\else
  \usepackage{cite}
\fi
\ifCLASSINFOpdf
\else
\fi
\begin{document}
%
\title{DocFace+: ID Document to Selfie\textsuperscript{*}\thanks{\textsuperscript{*}~Technically, the word ``selfies'' refers to self-captured photos from mobile phones. But here, we define ``selfies'' as any self-captured live face photos, including those from mobile phones and kiosks.} Matching}
%

\author{Yichun~Shi,~\IEEEmembership{Student Member,~IEEE,}
     and Anil K. Jain,~\IEEEmembership{Life Fellow,~IEEE}

\IEEEcompsocitemizethanks{\IEEEcompsocthanksitem Y. Shi and A. K. Jain are with the Department of Computer Science and Engineering, Michigan State University, East Lansing, MI, 48824.\protect\\
E-mail: shiyichu@msu.edu, jain@cse.msu.edu}
}

%
%

\markboth{}
{Shell \MakeLowercase{\textit{et al.}}: Bare Demo of IEEEtran.cls for Biometrics Council Journals}
%



\IEEEtitleabstractindextext{%

\begin{abstract}
   Numerous activities in our daily life require us to verify who we are by showing our ID documents containing face images, such as passports and driver licenses, to human operators. However, this process is slow, labor intensive and unreliable. As such, an automated system for matching ID document photos to live face images (selfies) in real time and with high accuracy is required. In this paper, we propose DocFace+ to meet this objective. We first show that gradient-based optimization methods converge slowly (due to the underfitting of classifier weights) when many classes have very few samples, a characteristic of existing ID-selfie datasets. To overcome this shortcoming, we propose a method, called dynamic weight imprinting (DWI), to update the classifier weights, which allows faster convergence and more generalizable representations. Next, a pair of sibling networks with partially shared parameters are trained to learn a unified face representation with domain-specific parameters. Cross-validation on an ID-selfie dataset shows that while a publicly available general face matcher (SphereFace) only achieves a True Accept Rate (TAR) of $59.29\pm 1.55\%$ at a False Accept Rate (FAR) of $0.1\%$ on the problem, DocFace+ improves the TAR to $97.51\pm 0.40\%$.
\end{abstract}

\begin{IEEEkeywords}
ID-selfie face matching, face recognition, face verification, access control, document photo, selfies
\end{IEEEkeywords}}

\maketitle

\IEEEdisplaynontitleabstractindextext

%
\IEEEpeerreviewmaketitle

\IEEEraisesectionheading{\section{Introduction}\label{sec:introduction}}

%
%
%
%

\begin{figure}[t]
\centering
\includegraphics[width=\linewidth]{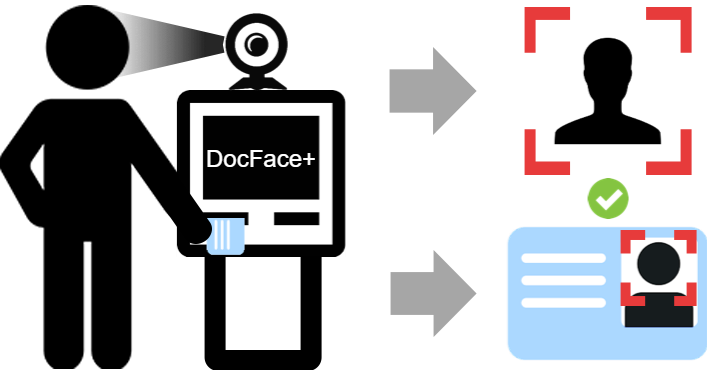}
\caption{An application scenario of the DocFace+ system. The kiosk scans the ID document photos or reads the photo from the embedded chip and the camera takes another photo of the holder's live face (selfie). By comparing the two photos, the system decides whether the holder is indeed the owner of the ID document.}
\label{fig:application}
\end{figure}

\IEEEPARstart{I}{dentity} verification plays an important role in our daily lives. For example, access control, physical security and international border crossing require us to verify our access (security) level and our identities. 
A practical and common approach to this problem involves comparing an individual's live face to the face image found in his/her ID document.
For example, immigration and customs officials look at the passport photo to confirm a traveler's identity.
Clerks at supermarkets in the United States look at the customer's face and driver license to check his/her age when the customer is purchasing alcohol. Instances of ID document photo matching can be found in numerous scenarios. However, it is primarily conducted by humans manually, which is time consuming, costly, and prone to operator errors. A study pertaining to the passport officers in Sydney, Australia, shows that even the trained officers perform poorly in matching unfamiliar faces to passport photos, with a $14\%$ false acceptance rate~\cite{white2014passport}. Therefore, an accurate and automated system for efficient matching of ID document photos to selfies\textsuperscript{*} is required. In addition, automated ID-selfie matching systems also enable remote authentication applications that are otherwise not feasible, such as onboarding new customers in a mobile app (by verifying their identities for account creation), or account recovery in the case of forgotten passwords. One application scenario of an ID-selfie matching system (DocFace+) is illustrated in Figure~\ref{fig:application}.

\begin{figure*}
\center
\subfigure[SmartGate (Australia)~\cite{gates_au}]{
    \includegraphics[width=0.22\linewidth]{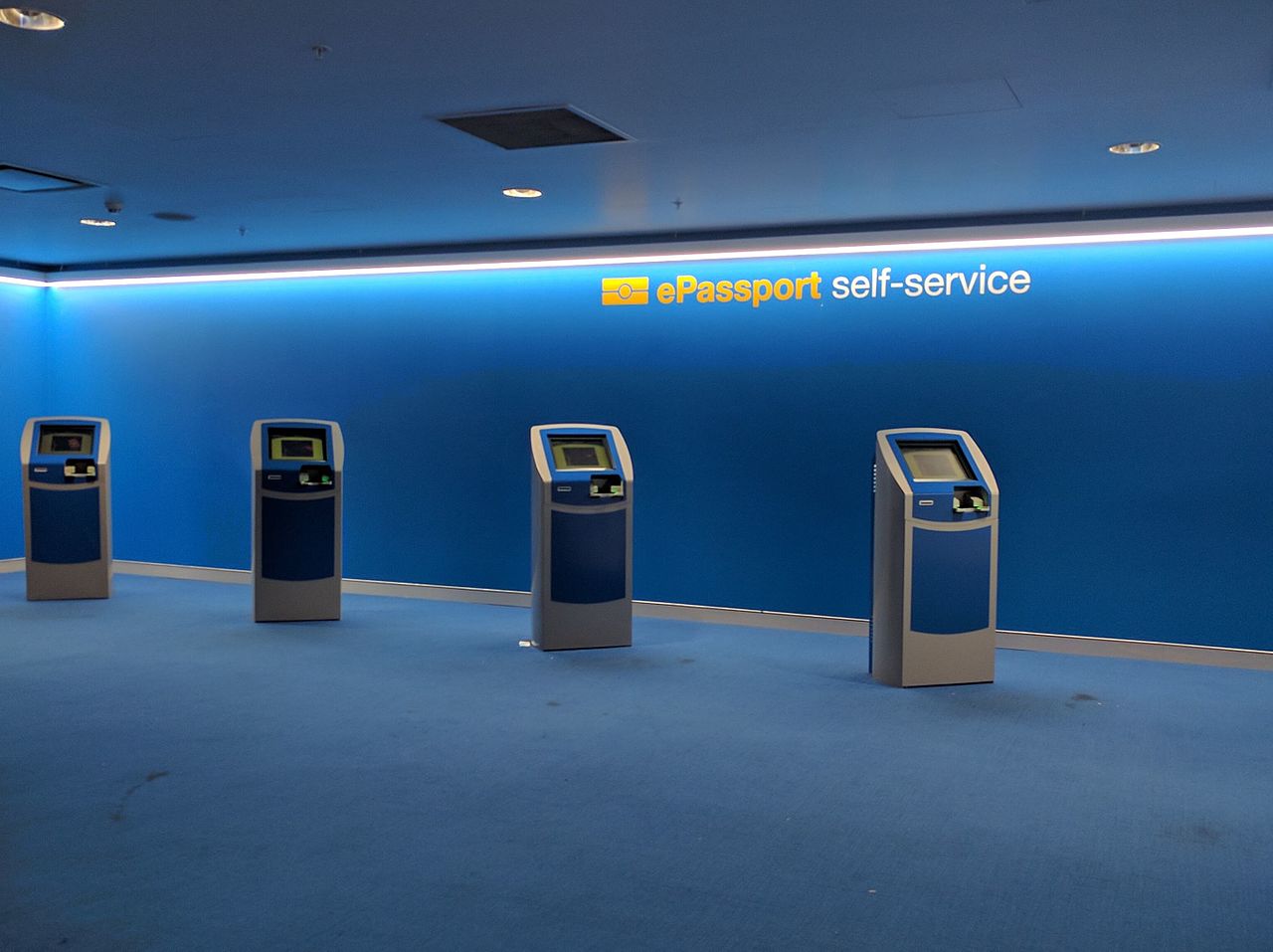}
}
\subfigure[ePassport gates (UK)~\cite{gates_uk}]{
    \includegraphics[width=0.24\linewidth]{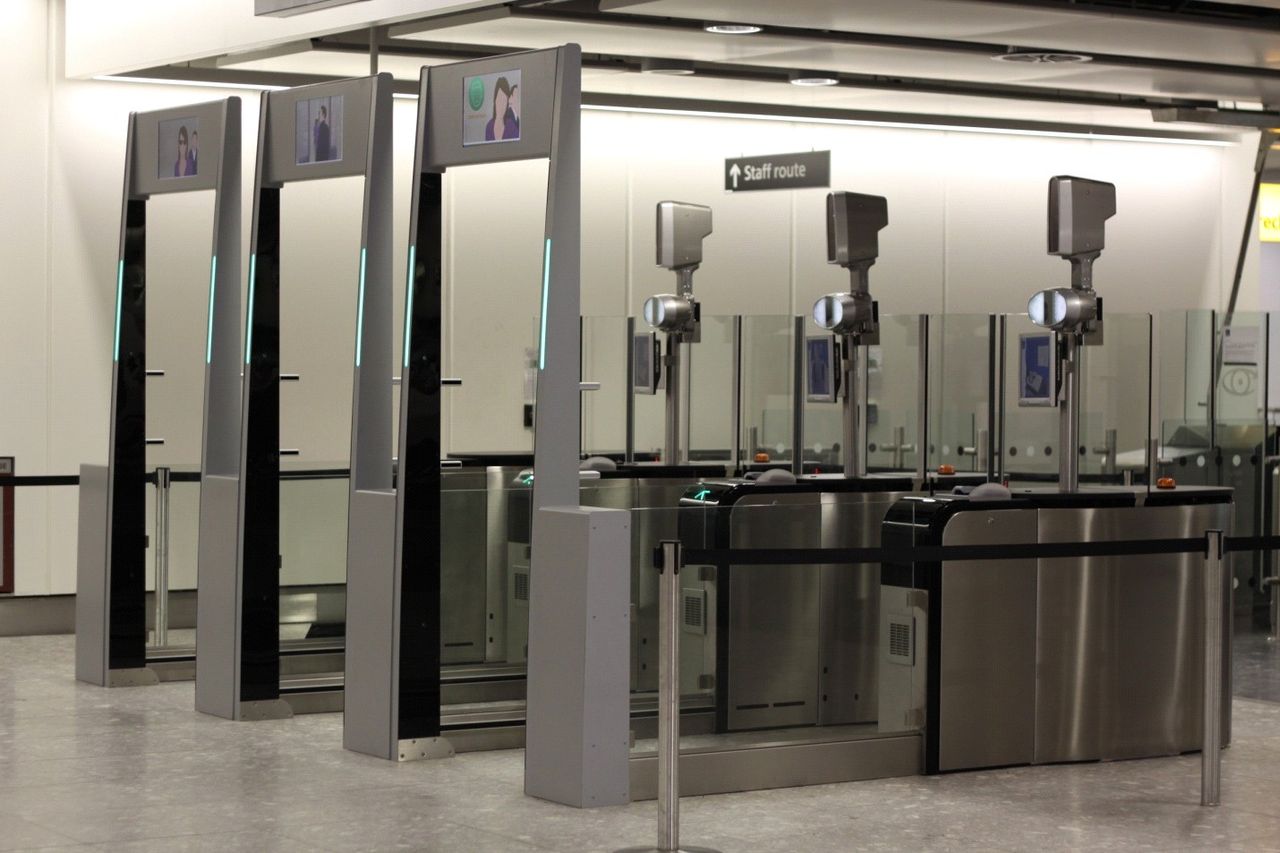}
}
\subfigure[Automated Passport Control (US)~\cite{gates_us}]{
    \includegraphics[width=0.24\linewidth]{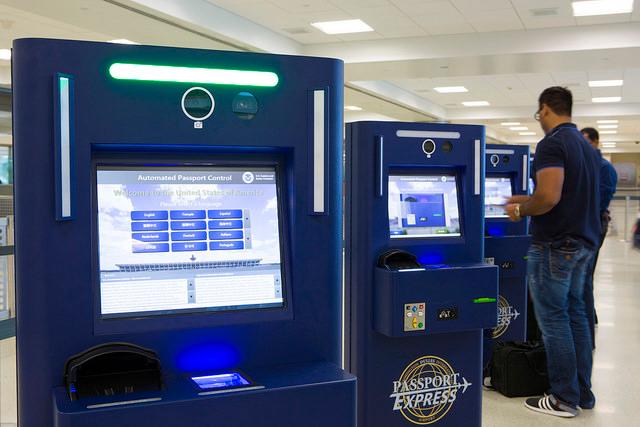}
}
\subfigure[ID card gates (China) ~\cite{gates_cn}]{
    \includegraphics[width=0.22\linewidth]{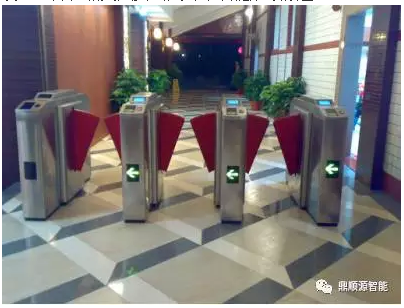}
}
\caption{Examples of automatic ID document photo matching systems at international borders.}
\label{fig:gates}
\end{figure*}
A number of automated ID-selfie matching systems have been deployed at international borders. Deployed in 2007, SmartGate~\cite{gates_au} in Australia (See Figure~\ref{fig:gates}) is the earliest of its kind. Due to an increasing number of travelers to Australia, the Australian government introduced SmartGate at most of its international airports as an electronic passport check for ePassport holders. To use the SmartGate, travelers only need to let a machine read their ePassport chips containing their digital photos and then capture their face images using a camera mounted at the SmartGate. After verifying a traveler's identity by face comparison, the gate is automatically opened for the traveler to enter Australia. Similar machines have also been installed in the UK (ePassport gates) ~\cite{gates_uk}, USA (US Automated Passport Control)~\cite{gates_us} and other countries. In China, such verification systems have been deployed in variouis locations, including train stations, for matching Chinese ID cards with live faces~\cite{gates_cn}. 
In addition to international border control, some businesses~\cite{netverify}~\cite{mitek} are utilizing face recognition solutions to ID document verification for online services.

\begin{figure}[t]
\center
\subfigure[General face matching]{
    \includegraphics[width=\linewidth]{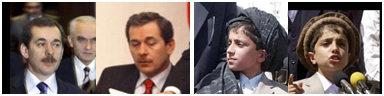}
}
\subfigure[ID-selfie matching]{
    \includegraphics[width=\linewidth]{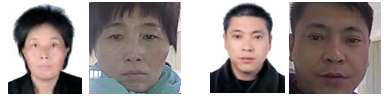}
}
\caption{Example images from (a) LFW dataset~\cite{LFWTech} and (b) ID-selfie dataset. Each row shows two pairs from each dataset, respectively. Compared with the general unconstrained face recognition shown in (a), ID Document photo matching in (b) does not need to consider large pose variations. Instead, it involves some other challenges such as aging and information loss via image compression.}
\label{fig:examples}
\end{figure}

The problem of ID-selfie matching poses numerous challenges that are different from general face recognition. For typical unconstrained face recognition tasks, the main challenges are due to pose, illumination and expression (PIE) variations. Instead, in ID-selfie matching, we are comparing a scanned or digital document photo to a digital camera photo of a live face. Assuming that the user is cooperative, both of the images are captured under constrained conditions and large PIE variations would not be present. Instead, (1) the low quality of document photos due to image compression\footnote{Most chips in e-Passports have a memory ranging from 8KB to 30KB; the face images need to be compressed to be stored in the chip. See \url{https://www.readid.com/blog/face-images-in-ePassports}} and (2) the large time gap between the document issue date and the verification date remain as the primary difficulties. See Figure~\ref{fig:examples}. In addition, since state-of-the-art face recognition systems are based on deep networks, another issue faced in our problem is the lack of a large training dataset (pairs of ID photos and selfies). 

In spite of numerous applications and associated challenges, there is a paucity of research on ID-selfie matching. Most of the published studies are now dated~\cite{starovoitov2000matching}\cite{bourlai2009matching}\cite{bourlai2011restoring}\cite{starovoitov2002three}. It is important to note that face recognition technology has made tremendous strides in the past five years, mainly due to the availability of large face datasets and the progress in deep neural network architectures. Hence, the earlier published results on ID-selfie matching are now obsolete. To the best of our knowledge, our prior work~\cite{shi2018docface} is the first to investigate the application of deep CNN to this problem, concurrent with Zhu et al.~\cite{zhu2018large}.

In this paper, we first briefly review existing studies on the ID-selfie matching problem and other researches related to our work. We then extend our prior work of DocFace~\cite{shi2018docface} to a more advanced method, DocFace+, for building ID-selfie matching systems. We use a large dataset of Chinese Identity Cards with corresponding selfies\footnote{This dataset consists of $53,591$ different ID-selfie pairs captured at different locations and at different times. Due to privacy reasons, we cannot release this data in the public domain. To our knowledge, there is no such dataset available in public domain.} to develop the system and to evaluate the performance of (i) two Commercial-Off-The-Shelf (COTS) face matchers, (ii) open-source deep network face matchers, and (iii) the proposed method. We also compare the proposed system on the open benchmark established by Zhu et al~\cite{zhu2018large}. The contributions of the paper are summarized below:

\begin{itemize}
    \item A new optimization method for classification-based embedding learning on shallow datasets\footnote{Face recognition datasets are often described in terms of breadth and depth\cite{cao2018vggface2}, where breadth refers to the number of classes and depth means the average number of samples per class.}.
    \item A new recognition system containing a pair of partially shared networks for learning unified representations from ID-selfie pairs.
    \item An evaluation of COTS and public-domain face matchers showing ID-selfie matching is a non-trivial problem with different challenges from general face matching. 
    \item An open-source face matcher\footnote{The source code is available at \url{https://github.com/seasonSH/DocFace}}, namely DocFace+, for ID-selfie matching, which significantly improves the performance of state-of-the-art general face matchers. Our experiment results show that while the publicly available CNN matcher (SphereFace) only achieves a True Accept Rate (TAR) of $59.29\pm 1.55\%$ at False Accept Rate (FAR) of $0.1\%$ on the problem, DocFace+ improves the TAR to $97.51\pm 0.40\%$.
\end{itemize}

\section{Related Works}
 
\subsection{ID Document Photo Matching}
To the best of our knowledge, the first study on ID-selfie matching is attributed to Starovoitov et al.~\cite{starovoitov2000matching}~\cite{starovoitov2002three}. Assuming all face images are frontal faces without large expression variations, the authors first localize the eyes with Hough Transform. Based on eye locations, the face region is cropped and gradient maps are computed as feature maps. The algorithm is similar to a general constrained face matcher, except it is developed for a document photo dataset. Bourlai et al.~\cite{bourlai2009matching}\cite{bourlai2011restoring} considered ID-selfie matching as a comparison between degraded face images, i.e. scanned document photos, and a high quality live face images. 
To eliminate the degradation caused by scanning, Bourlai et al. inserted an image restoration phase before comparing the photos using a general face matcher. In particular, they train a classifier to classify the degradation type for a given image, and then apply degradation-specific filters to restore the degraded images. Compared with their work on scanned documents, the document photos in our datasets are read from the chips embedded in the Chinese ID Cards. Additionally, our method is not designed for any specific degradation type and could be applied to any type of ID document photos. Concurrent with our prior work~\cite{shi2018docface}, Zhu et al.~\cite{zhu2018large} also worked on deep CNN-based ID-selfie matching systems. With 2.5M ID-selfie pairs, also from a private Chinese ID card dataset, they formulated it as a bisample learning problem and proposed to train the network in three stages: (1) pre-learning (classification) on general face datasets, (2) transfer learning (verification) and (3) fine-grained learning (classification). Our work, on the other hand, proposes a special optimization method to address the slow convergence problem of classification-based embedding learning methods on ID-selfie datasets, which does not require multi-stage training. Compared with our prior work~\cite{shi2018docface}, the differences of this version are as follows: (1) a larger ID-selfie dataset (over 50,000 papers), which is a combination of the two small private datasets in~\cite{shi2018docface} and another larger ID-selfie dataset, (2) a different loss function, namely DIAM-Softmax, to learn the face representation, (3) more comprehensive experiments to analyze the effect of each module and (4) an evaluation of the proposed system as well as other face matchers on a new ID-selfie benchmark, Public IvS, released by Zhu et al. in~\cite{zhu2018large}.

\begin{figure*}[t]
\centering
\subfigure[MS-Celeb-1M]{
    \includegraphics[width=0.31\linewidth]{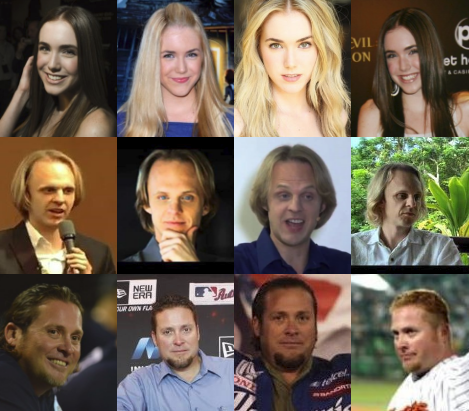}
}
\subfigure[Private ID-selfie]{
    \includegraphics[width=0.31\linewidth]{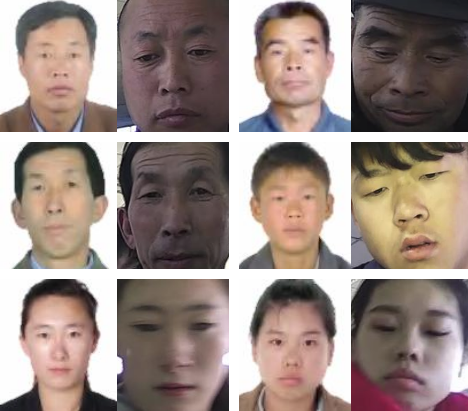}
}
\subfigure[Public IvS]{
    \includegraphics[width=0.31\linewidth]{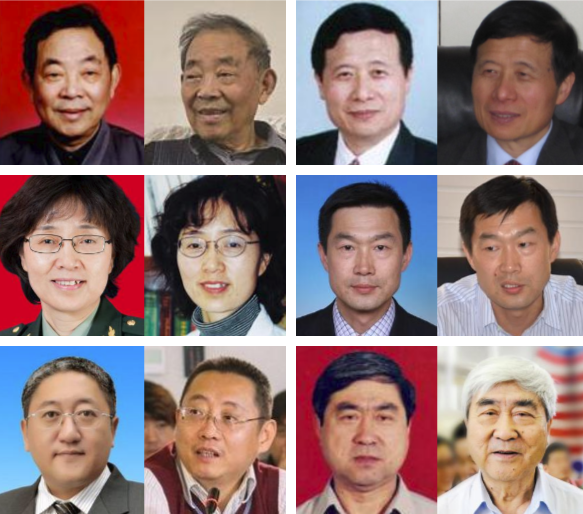}
}
\caption{Example images in each dataset. The left image in each pair in (b) and (c) is the ID photo and on its right is one of its corresponding selfies.}
\label{fig:datasets}
\end{figure*}

\subsection{Deep Face Recognition}
Since the success of deep neural networks in the ImageNet competition~\cite{krizhevsky2012imagenet}, virtually all of the ongoing research in face recognition now utilize deep neural networks to learn face representation~\cite{taigman2014deepface}~\cite{deepid2}~\cite{schroff2015facenet}~\cite{liu2017sphereface}~\cite{hasnat2017deepvisage}~\cite{wang2018additive}. Taigman et al.~\cite{taigman2014deepface} first proposed to apply deep neural networks to learning face representation. They designed an 8-layer convolutional neural network and trained it with a Softmax loss function, a standard loss function for classification problems. They used the outputs of the bottleneck layer as face representation and achieved state-of-the-art performance at that time in 2014. Considering that Softmax loss only encourages large inter-class variations but does not constrain intra-class variations, Sun et al.~\cite{deepid2} later proposed to train networks with both a classification signal (Softmax loss) and a metric learning signal (Contrastive loss). About the same time, Schroff et al.~\cite{schroff2015facenet} proposed a metric learning loss function, named triplet loss, boosting the state-of-the-art performance on the standard LFW protocol~\cite{LFWTech}. Liu et al.~\cite{liu2017sphereface} first bridged the gap between classification loss functions and metric learning methods with the Angular-Softmax (A-Softmax) loss function, a modified Softmax loss that classifies samples based on angular distances. Wang et al.~\cite{wang2018additive} recently proposed the Additive Margin Softmax (AM-Softmax), which learns to increase the angular discriminability with an additive margin and is shown to be more robust than A-Softmax.


\subsection{Heterogeneous Face Recognition}
Heterogeneous face recognition (HFR) is an emerging topic that has become popular in the past few years~\cite{li2009encyclopedia}. It usually refers to face recognition between two different modalities, including visible spectrum images (VIS), near infrared images (NIR)~\cite{li2007illumination}, thermal infrared images~\cite{choiHYD12}, composite sketches~\cite{tang2002face}, etc. ID-selfie matching can be considered to be a special case of HFR since the images to be matched come from two different sources and thus need possibly different modeling approaches. Therefore, the techniques used in HFR could be potentially helpful for the ID-selfie problem. Most methods for HFR can be categorized into two types: synthesis-based methods and discriminant feature-based methods. Synthesis-based methods aim to transform the images from one modality into another so that general intra-modality face recognition systems can be applied~\cite{tang2002face}\cite{wang2008face}\cite{liu2005nonlinear}\cite{gao2008face}. On the contrary, discriminant feature-based methods either manually design a modality-invariant visual descriptor or learn a set of features from the training data so that images from different modalities can be mapped into a shared feature space~\cite{liao2009heterogeneous}\cite{klare2010heterogeneous}\cite{klare2013heterogeneous}. Recent studies on HFR have focused on utilizing deep neural networks to learn such modality-invariant features. In~\cite{liu2016transferring}, Liu et al. first proposed to apply deep VIS face features to VIS-NIR matching via transfer learning where their network is fine-tuned on VIS-NIR dataset with a modified triplet loss. He et al. proposed~\cite{he2017learning} to use a shared convolutional neural network to map VIS and NIR images into three feature vectors: one set of shared features and two modality-specific feature sets, which are then concatenated and trained with three Softmax loss functions. Wu et al.~\cite{wu2017coupled} proposed to learn a unified feature space with two correlated modality-specific Softmax loss functions. The weights of the two Softmax loss are regularized by a trace norm and a block-diagonal prior to encourage correlation between representations from different modalities and to avoid overfitting.

\subsection{Low-shot Learning}
Another field related to our work is the low-shot learning problem. In low-shot learning ~\cite{koch2015siamese}\cite{vinyals2016matching}\cite{snell2017prototypical}, a model is trained in such a way that it is able to generalize to unseen classes, which may have only a few samples. There are two training phases in low-shot learning: the model, or learner, is first trained on a larger classification dataset, and then in testing, a few labeled samples of new classes are given and the model is required to learn a new classifier given these classes. The adjective ``low-shot'' refers to a small number of images per class. This problem has been receiving growing interest from the machine learning community because humans are very good at adapting to new types of objects (classes) while conventional deep learning methods require abundant samples for discriminating a specific class from others. This is related to the ID-selfie problem since most of the identities only have a few samples, resulting in a shallow dataset. Many methods have been proposed for low-shot learning problem. Koch et al.~\cite{koch2015siamese} proposed a simple yet effective approach by learning metrics via siamese network~\cite{chopra2005learning} for one-shot recognition. Vinyals et al.~\cite{vinyals2016matching} proposed the Matching Net where they simulate the testing scenario in the training phase by learning low-shot recognition in mini-batches. This idea was then generalized as \textit{meta-learning}, where an extra meta-learner can learn how to optimize or produce new classifiers~\cite{ravi2016optimization}\cite{bertinetto2016learning}. The Prototypical Network by Snell et al.~\cite{snell2017prototypical} is more relevant to our work. They proposed to learn a network such that prototypes, i.e. average feature vector of an unseen class, can be used for classification. Qi et al.~\cite{qi2018lowshot}, based on the idea of the prototypes, or proxies~\cite{movshovitz2017no}, proposed to imprint the weights of a new classifier with extracted features. We note that their work differs from ours as they utilized the imprinted weights simply as initialization while we use weight imprinting as an optimization method throughout training by dynamically imprinting the weights.

\section{Datasets}
In this section, we briefly introduce the datasets that are used in this paper. Some example images of the datasets are shown in Figure~\ref{fig:datasets}. As stated earlier, due to privacy issues, we cannot release the Private ID-selfie dataset. But by comparing our results with public face matchers, we believe it is sufficient to show the difficulty of the problem and advantages of the proposed method.

\subsection{MS-Celeb-1M}
The MS-Celeb-1M dataset~\cite{guo2016msceleb} is a public domain face dataset facilitating training of deep networks for face recognition. It contains $8,456,240$ face images of $99,892$ subjects (mostly celebrities) downloaded from internet. In our transfer learning framework, it is used to train a base network.
Because the dataset is known to have many mislabels, we use a cleaned version\footnote{\url{https://github.com/AlfredXiangWu/face_verification_experiment}.} of MS-Celeb-1M with $5,041,527$ images of $98,687$ subjects. Some example images from this dataset are shown in Figure~\ref{fig:datasets}(a).

\subsection{Private ID-selfie}
\label{sec:dataset_A}
During the experiments, we use a private dataset to develop and evaluate our ID-selfie matching system. It is a combination of a larger ID-selfie dataset and two other smaller datasets used in our prior work~\cite{shi2018docface}. The dataset contains $116,914$ images of $53,591$ identities in all. Each identity has only one ID card photo. A subset of $53,054$ identities have only one selfie while the other $537$ have multiple selfies. The ID card photos are read from chips in the Chinese Resident Identity Cards\footnote{The second-generation Chinese ID cards, which were launched in 2004 and have completely replaced the first generation in 2014, contain a IC chip. The chip stores a compressed face photo of the owner. See more at \url{https://en.wikipedia.org/wiki/Resident_Identity_Card}}. In experiments, we will conduct 5-fold cross-validation on this dataset to evaluate the efficacy of each part of our method. Some example pairs from this dataset are shown in Figure~\ref{fig:datasets}(b).

\subsection{Public IvS}
\label{sec:dataset:public}
Public IvS is a dataset released by Zhu et al.~\cite{zhu2018large} for evaluation of ID-selfie matching systems. The dataset is constructed by collecting ID photos and live face photos of Chinese personalities from Internet. The dataset contains $1,262$ identities and $5,503$ images in total. Each identity has one ID photo and 1 to 10 selfies. It is not strictly an ID-selfie dataset since its ID photos are not from real ID cards but are simulated with highly constrained frontal photos. The results on this dataset were shown to be consistent with real-world ID-selfie datasets~\cite{zhu2018large}. 
Some example pairs of this dataset are shown in Figure~\ref{fig:datasets}(c).

\begin{figure}[t]
\centering
\includegraphics[width=\linewidth]{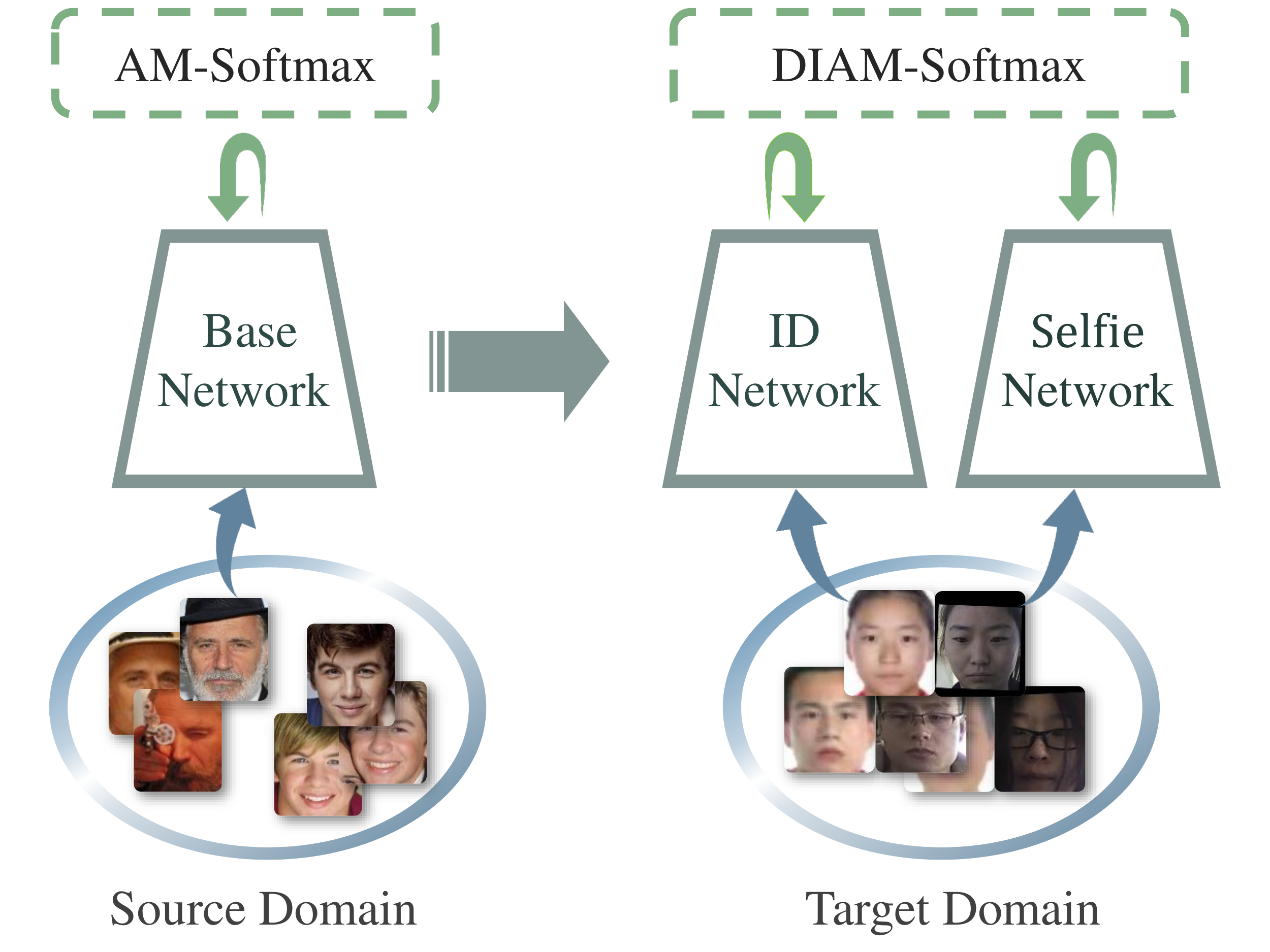}
\caption{Work flow of the proposed method. We first train a base model on a large scale unconstrained face dataset. Then, the parameters are transferred to a pair of sibling networks, who have shared high-level modules. Random ID-selfie pairs are sampled from different classes to train the networks. The proposed DIAM-Softmax is used to learn a shared feature space for both domains of ID and selfie.}

\label{fig:overview}
\end{figure}

\section{Methodology}
\subsection{Overview}
In our work, we first train a network as \emph{base model} on a large-scale unconstrained face dataset, i.e. MS-Celeb 1M and then transfer its features to our target domain of ID-selfie pairs. To ensure the performance of transfer learning, we utilize the popular Face-ResNet architecture~\cite{hasnat2017deepvisage} to build the convolutional neural network. For training the base model, we adopt state-of-the-art \textit{Additive Margin Softmax} (AM-Softmax) loss function~\cite{wang2018additive}\cite{wang2018cosface}. Then we propose a novel optimization method called dynamic weight imprinting (DWI) to update the weight matrix in AM-softmax when training on the ID-selfie dataset. A pair of sibling networks is proposed for learning domain-specific features of IDs and selfies, respectively, with shared high-level parameters. An overview of the work flow is shown in Figure~\ref{fig:overview}.


\subsection{Original AM-Softmax}
We use the original \textit{Additive Margin Softmax} (AM-Softmax) loss function~\cite{wang2018additive}\cite{wang2018cosface} for training the base model. Here, we give a short review to gain more understanding of this loss function before we introduce our modifications in the next section. Similar to Angular Softmax~\cite{liu2017sphereface} and L2-Softmax~\cite{ranjan2017l2}, AM-Softmax is a classification-based loss function for embedding learning, which aims to maximize inter-subject separation and to minimize intra-subject variations. Let $X^s=\{(x_i, y_i)|i=1,2,3,\cdots, N\}$ be our training dataset and $\mcF:\IR^{h\times w\times c} \rightarrow \IR^{d}$ be a feature extraction network, where $x_i\in \IR^{h\times w\times c}$ is a face image, $y_i$ is the label and $h$,$w$,$c$ are the height, width and number of channels of the input images, respectively. $N$ is the number of training images and $d$ is the number of feature dimensions. For a training sample $x_i$ in a mini-batch, the loss function is given by:

\begin{equation}
    \EL ={-\log p^{(i)}_{y_i}}
\label{eq:loss_additive}
\end{equation}
where
\begin{gather*}
    p^{(i)}_{j} = \frac{ \exp(a^{(i)}_{j}) }{\sum_{k}{\exp(a^{(i)}_{k})}}\\
    a^{(i)}_{j} = \begin{cases}
      s\bwj^T\bfi - m,  & \text{if}\ j=y_i \\
      s\bwj^T\bfi,      & \text{otherwise}
    \end{cases}\\
    \bwj=\frac{\bwj^*}{\|\bwj^*\|_2} \\
    \bfi=\frac{\mcF(x_i)}{\|\mcF(x_i)\|_2}.
\end{gather*}
Here $\bwj^*\in\IR^{d}$ is the weight vector for $j^{th}$ class, $m$ is a hyper-parameter for controlling the margin and $s$ is a scale parameter. Notice that this formula is a little different from the original AM-Softmax~\cite{wang2018additive} in the sense that the margin $m$ is not multiplied by $s$, which allows us to automatically learn the parameter $s$~\cite{wang2017normface}. During training, the loss in Equation~(\ref{eq:loss_additive}) is averaged across all images in the mini-batch. The key difference between AM-Softmax and original Softmax is that both of the features $f_i$ and the weight vectors $w_i$ are normalized and lie in a spherical embedding space. Thus, instead of classifying samples based on inner products, AM-Softmax categorizes samples based on cosine similarity, the same metric used in the testing phase. This closes the gap between training and testing as well as the gap between classification learning and metric learning. The normalized weight vector $\bwj$ is considered to be an ``agent'' or ``proxy'' of the $j^{th}$ class, representing  the distribution of this class in the embedding space~\cite{wang2017normface}. During the optimization, classifier weights are also optimized with stochastic gradient descent (SGD). For the weights $\mathbf{w}_{y_i}$ of the ground-truth class, the gradient
\begin{equation}
\frac{\partial \EL}{\partial \mathbf{w}_{y_i}} = s(1-p^{(i)}_{y_i})\bfi
\label{eq:gradient_positive}
\end{equation}
serves as an attraction signal to pull $\mathbf{w}_{y_i}$ closer to $\bfi$. For other classes $j\neq y_i$, the gradient
\begin{equation}
\frac{\partial \EL}{\partial \bwj} = -sp^{(i)}_{j}\bfi
\label{eq:gradient_negative}
\end{equation}
provides a repulsion signal to push $\bwj$ away from $\bfi$. Compared with metric learning methods~\cite{chopra2005learning}~\cite{schroff2015facenet}, the normalized weights allows AM-Softmax to capture global distributions of different classes in the embedding space and leads to a faster convergence and better performance.

\begin{figure}[t]
\centering
\includegraphics[width=\linewidth]{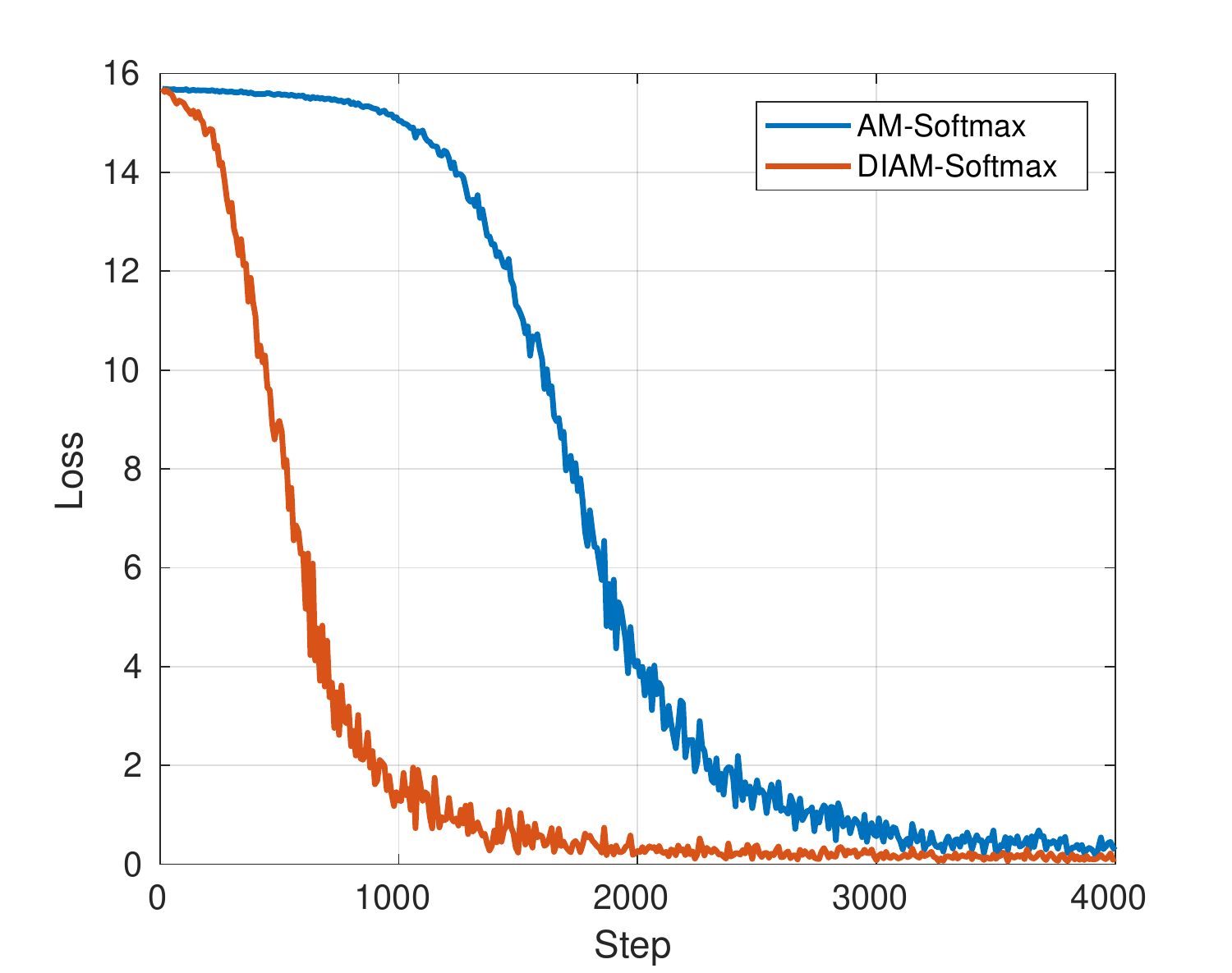}
\caption{Training loss of AM-Softmax compared with DIAM-Softmax. DIAM-Softmax shares the same formula as AM-Softmax, but its weights are updated with the proposed DWI instead of SGD.}
\label{fig:Loss}
\end{figure}

\begin{figure}[t]
\centering
\subfigure[AM-Softmax]{
    \includegraphics[width=0.46\linewidth]{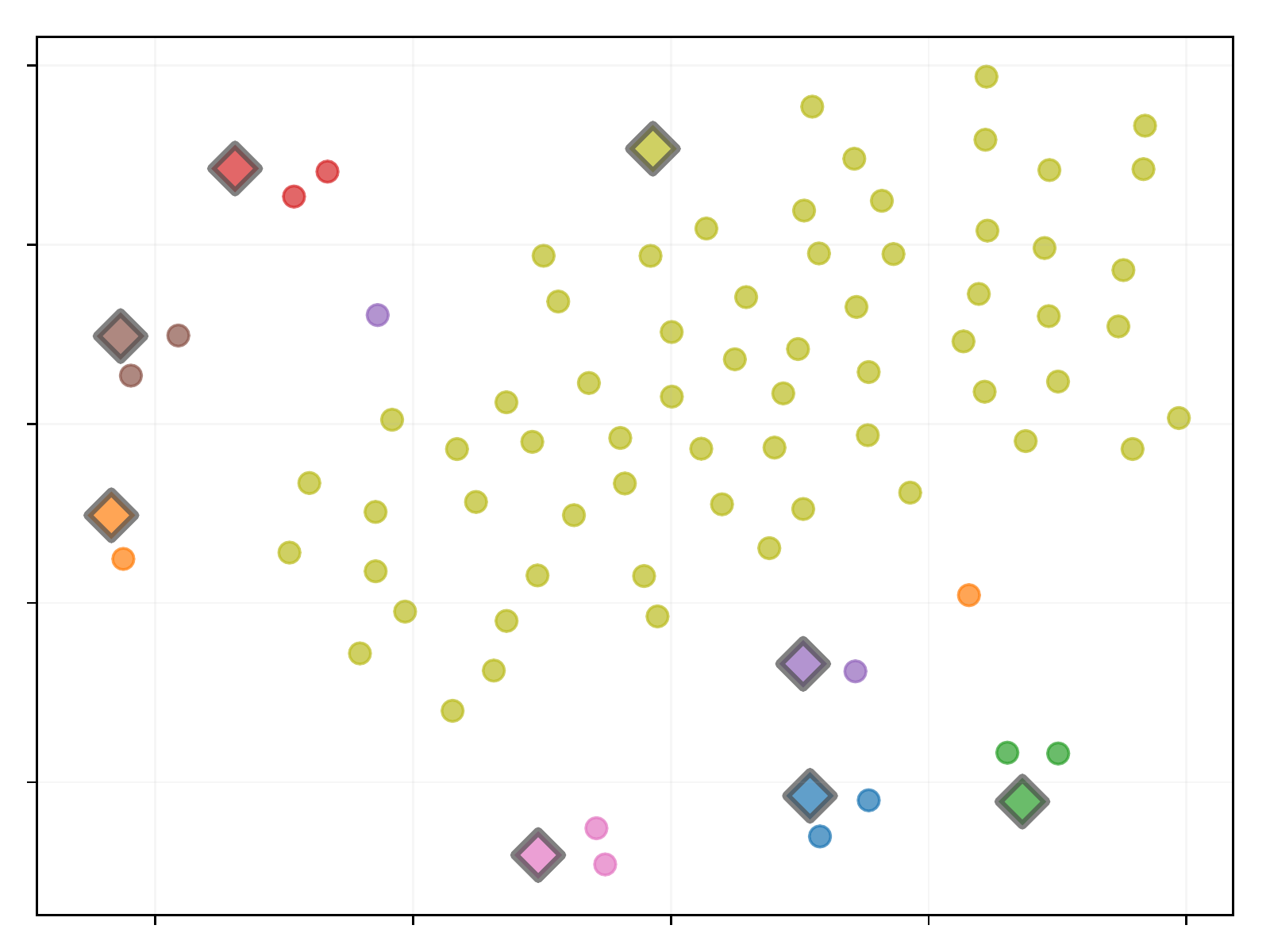}
}
\subfigure[DIAM-Softmax]{
    \includegraphics[width=0.46\linewidth]{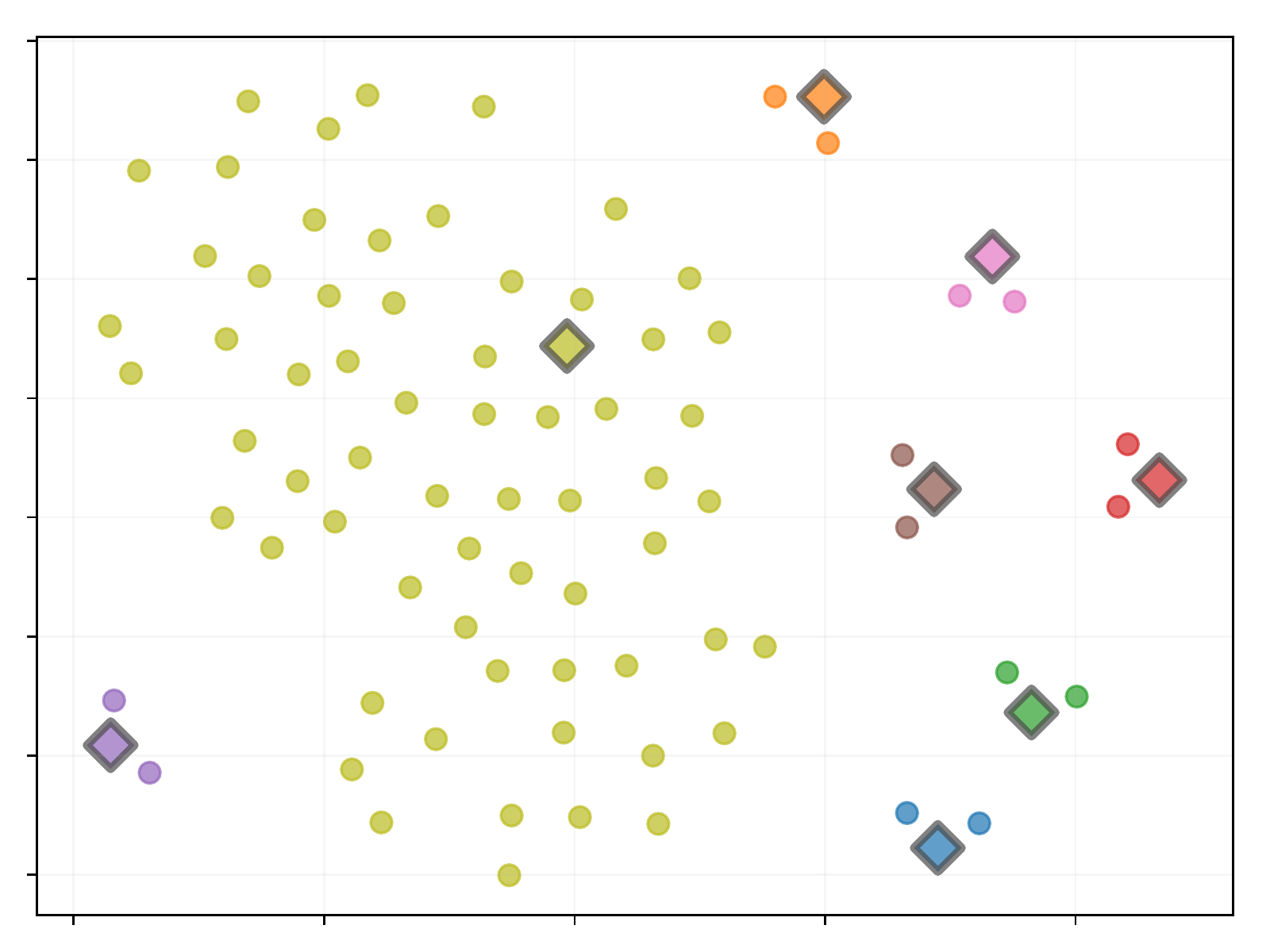}
}
\caption{Visualization of $8$ random classes in the training dataset. The dimensionality of the original embedding space is $512$ and is reduced to $2$ with t-SNE~\cite{maaten2008visualizing}. The circles are training samples and the diamonds are normalized classifier weights. In original AM-Softmax, many weight vectors shift from their corresponding distributions when the dataset is shallow and update signals are sparse, leading to a low convergence speed.}
\label{fig:embedding}
\end{figure}

\subsection{Dynamic Weight Imprinting}
\label{sec:weight_imprinting}
In spite of the success of AM-Softmax and other classification-based embedding learning loss functions on general face recognition~\cite{ranjan2017l2}\cite{wang2017normface}\cite{liu2017sphereface}, we found them to be less competitive for transfer learning on the ID-selfie dataset~\cite{shi2018docface}. In fact, it is often the case that they converge very slowly and get stuck at a bad local minimum. As shown in Figure~\ref{fig:Loss}, the original AM-Softmax does not start to converge after several epochs\footnote{In this example, each epoch is approximately $300$ steps.}. To gain more insight into the problem, we visualize the embeddings via dimensionality reduction. We extract the features of samples as well as the normalized weight vectors of $8$ classes\footnote{Seven classes are randomly selected and one class is chosen to have more data. Similar to most classes, the randomly selected seven classes have only two images (one ID and one selfie).} in the training dataset and reduce their dimensionality from $512$ to $2$ using t-SNE~\cite{maaten2008visualizing}. The visualization is shown in Figure~\ref{fig:embedding}(a). Noticeably, many weights are shifted from the distribution of the corresponding class even after convergence. Although the shift of weights is expected in the original loss function because of repulsion signals~\cite{wang2017normface} and does not harm the convergence on general face datasets, this shift becomes large and harmful on shallow datasets, where there are a large number of classes with only a few samples for most classes. Rather than repulsion from other classes, this larger shift is mainly caused by the optimization method. Since SGD updates the network with mini-batches, in a two-shot case, each weight vector will receive attraction signals only twice per epoch. After being multiplied by the learning rate, these sparse attraction signals make little difference to the classifier weights. Thus instead of overfitting, this sparseness of signals from SGD causes the underfitting of the classifier weights in the last fully connected layer, who shift from the feature distribution and lead to the slow convergence. 

\begin{table}[t]
\footnotesize
\begin{center}
\caption{The mean (and s.d. of) performance of AM-Softmax and DIAM-Softmax on the Private ID-selfie dataset based on 5-fold cross-validation.}
\label{tab:sgd_dwi}
\begin{tabularx}{\linewidth}{Xccc}
\toprule
Method   & \multicolumn{3}{c}{True Accept Rate (\%)} \\ \cmidrule(lr){2-4}
                                     & FAR=$0.001\%$     & FAR=$0.01\%$  & FAR=$0.1\%$  \\
\midrule
AM-Softmax                                  & $91.65\pm 1.19$   & $95.13\pm 0.72$       & $97.08\pm 0.45$   \\
AM-Softmax (2x steps)                       & $92.53\pm 1.09$   & $95.57\pm 0.57$       & $97.23\pm 0.42$   \\
DIAM-Softmax                                & $\mathbf{93.16\pm 0.85}$ & $\mathbf{95.95\pm 0.54}$ & $\mathbf{97.51\pm 0.40}$ \\
\bottomrule
\end{tabularx}
\end{center}
\end{table}

Based on the above observations, we propose a different optimization method for the weights in classification-based embedding learning loss functions. The main idea is to update the weights based on sample features to avoid underfitting of the classifier weights and accelerate the convergence. This idea of weight imprinting has been studied in the literature~\cite{qi2018lowshot}~\cite{zhu2018large}, but they only imprint the weights at the beginning of fine-tuning. Inspired by the center loss~\cite{wen2016discriminative}, we propose a dynamic weight imprinting (\textbf{DWI}) strategy for updating the weights:
\begin{equation}
\bwj = \frac{\bwj^*}{\|\bwj^*\|_2},
\label{eq:update_w}
\end{equation}
where
\begin{equation}
\bwj^* = (1-\alpha)\bwj + \alpha \bwjb
\label{eq:w_batch}
\end{equation}
Here $\bwjb$ is a target weight vector that is computed based on current mini-batch.
Notice that we only update the weights of classes whose samples are present in the current mini-batch and we store $\bwj$ rather than $\bwj^*$ as as variables. We consider three candidates for $\bwjb$ in our ID-selfie problem: (1) the feature of ID image, (2) the feature of selfie image and (3) the mean feature of ID and selfie images. The hyper-parameter $\alpha$ is the update rate. We are using this $\alpha$ here to consider a broader case where the weights are softly updated. In fact, as shown in Section~\ref{sec:exp:dwi}, $\alpha=1$ actually leads to the best performance, in which case the update formula can be simply written as:
\begin{equation}
\bwj = \frac{\bwjb}{\|\bwjb\|_2}\\
\label{eq:update_w_hard}
\end{equation}

Intuitively, DWI helps to accelerate the updating of weight vectors by utilizing sampled features and is invariant to the parameter settings of optimizers. Compared with gradient-based optimization, it only updates the weights based on genuine samples and does not consider repulsion from other classes. This may raise doubts on whether it could optimize the loss function in Equation~(\ref{eq:loss_additive}). However, as shown in Figure~\ref{fig:Loss} and Figure~\ref{fig:embedding}, empirically we found DWI is not only able to optimize the loss function, but it also helps the loss converge much faster by reducing the shift of weights. Furthermore, our cross-validation results on the Private ID-selfie dataset show that DWI is superior to SGD in terms of accuracy even when we train SGD for twice as long, until complete convergence. See Table~\ref{tab:sgd_dwi}.

Notice that DWI is not specially designed for AM-Softmax, but can be applied to other classification-based embedding learning loss functions as well, such as L2-Softmax~\cite{ranjan2017l2}. Although it is mainly an optimization method and it does not change the formula of the loss function, from another perspective, DWI also results in a new loss function, since different choices of classifier weights essentially poses
different learning objectives. Therefore, we name the method used in this paper, which combines DWI and AM-Softmax, as a new loss function, called Dynamically Imprinted AM-Softmax (\textbf{DIAM-Softmax}).

It is important to note here that DWI does not introduce any significant computational burden and the training speed is almost the same as before. In addition, since DWI only updates the weights of classes that are present in the mini-batch, it is naturally compatible with extremely wide datasets where the weight matrix of all classes is too large to be loaded and only a subset of weights can be sampled for training, as in~\cite{zhu2018large}. However, because of the data limitations, we do not further explore this idea in this paper.

\subsection{Domain Specific Modeling}
The ID-selfie matching problem can be regarded as an instance of heterogeneous face recognition (HFR)~\cite{klare2013heterogeneous}, since the face images come from two different sources. Thus, it is reasonable to expect that HFR methods could help with our problem. A common approach in HFR is to utilize two separate domain-specific models to map images from different sources into a unified feature space. Therefore, we use a pair of \textit{sibling networks} for ID images and selfie images, respectively, which share the same architecture but could have different parameters. Both of their features are transferred from the base model, i.e. they have the same initialization. Although this increases the model size, the inference speed will remain unchanged as each image is only fed into one of the sibling networks. The use of sibling networks allows domain-specific modeling, but more parameters could also lead to a higher risk of overfitting. Therefore, different from our prior work~\cite{shi2018docface}, we propose to constrain the high-level parameters of the sibling networks to be shared to avoid overfitting. In particular, we use a pair of Face-ResNet models with only the bottleneck layer shared. 

\subsection{Data Sampling}
When we apply classification-based loss functions to general face recognition, the mini-batches for training are usually constructed by sampling images uniformly at random. However, because our data is acquired from two different domains, such an image-level uniform sampling may not be the optimal choice. There is usually only one ID image in one class while there could be many more selfies, so sampling images uniformly could lead to a bias towards selfies and hence an insufficient modeling of the ID domain. Therefore, we propose to use a different sampling strategy to address the domain imbalance problem. In each iteration, $B/2$ classes are chosen uniformly at random, where $B$ is the batch size, and a random ID-selfie pair is sampled from each class to construct the mini-batch. Empirical results in Section~\ref{sec:exp:sampling} show that such a balanced sampling leads to a better performance compared with image-level uniform sampling.

\section{Experiments}
\label{sec:exp}
\subsection{Experimental Settings}
\label{sec:exp:settings}
We conduct all the experiments using Tensorflow library\footnote{\url{https://www.tensorflow.org}}. When training the base model with original AM-Softmax on MS-Celeb-1M, we use a batch size of $256$ and keep training for $280$K steps. We start with a learning rate of $0.1$, which is decreased to $0.01$, $0.001$ after $160$K and $240$K steps, respectively. When fine-tuning on the Private ID-selfie dataset, we use a batch size of $248$ and train the sibling networks for $4,000$ steps. We start with a lower learning rate of $0.01$ and decrease the learning rate to $0.001$ after $3,200$ steps. For both training stages, the feature networks are optimized by a Stochastic Gradient Descent (SGD) optimizer with a momentum of $0.9$ and a weight decay of $0.0005$. All the images are aligned via similarity transformation based on landmarks detected by MTCNN~\cite{zhang2016joint} and are resized to $96\times112$. We set margin parameters $m$ as $5.0$ in both stages. All the training and testing are run on a single Nvidia Geforce GTX 1080Ti GPU with $11$GB memory. The inference speed of our model on this GPU is $3$ms per image.

By utilizing the MS-Celeb-1M dataset and the AM-Softmax loss function in Equation~(\ref{eq:loss_additive}), our base model achieves $99.67\%$ accuracy on the standard verification protocol of LFW and a Verification Rate (VR) of $99.60\%$ at False Accept Rate (FAR) of $0.1\%$ on the BLUFR~\cite{liao2014benchmark} protocol.

In the following subsections, five-fold cross-validation is conducted for all experiments on Private ID-selfie dataset to evaluate the performance and robustness of the methods. The dataset is equally split into 5 partitions, and in each fold, one split is used for testing while the remaining are used for training. In particular, $42,873$ and $10,718$ identities are used for training and testing, respectively, in each fold. We use the whole Public IvS dataset for cross-dataset evaluation. Cosine similarity is used as comparison score for all experiments.

\subsection{Dynamic Weight Imprinting}
\label{sec:exp:dwi}
In this section, we want to compare how different weight imprinting strategies would affect the performance of DIAM-Softmax and find the best settings for the following experiments. We first compare the accuracy of using different update rates $\alpha$ and then evaluate different choices of update vector $\bwjb$.

\begin{figure}[t]
\centering
\includegraphics[width=\linewidth]{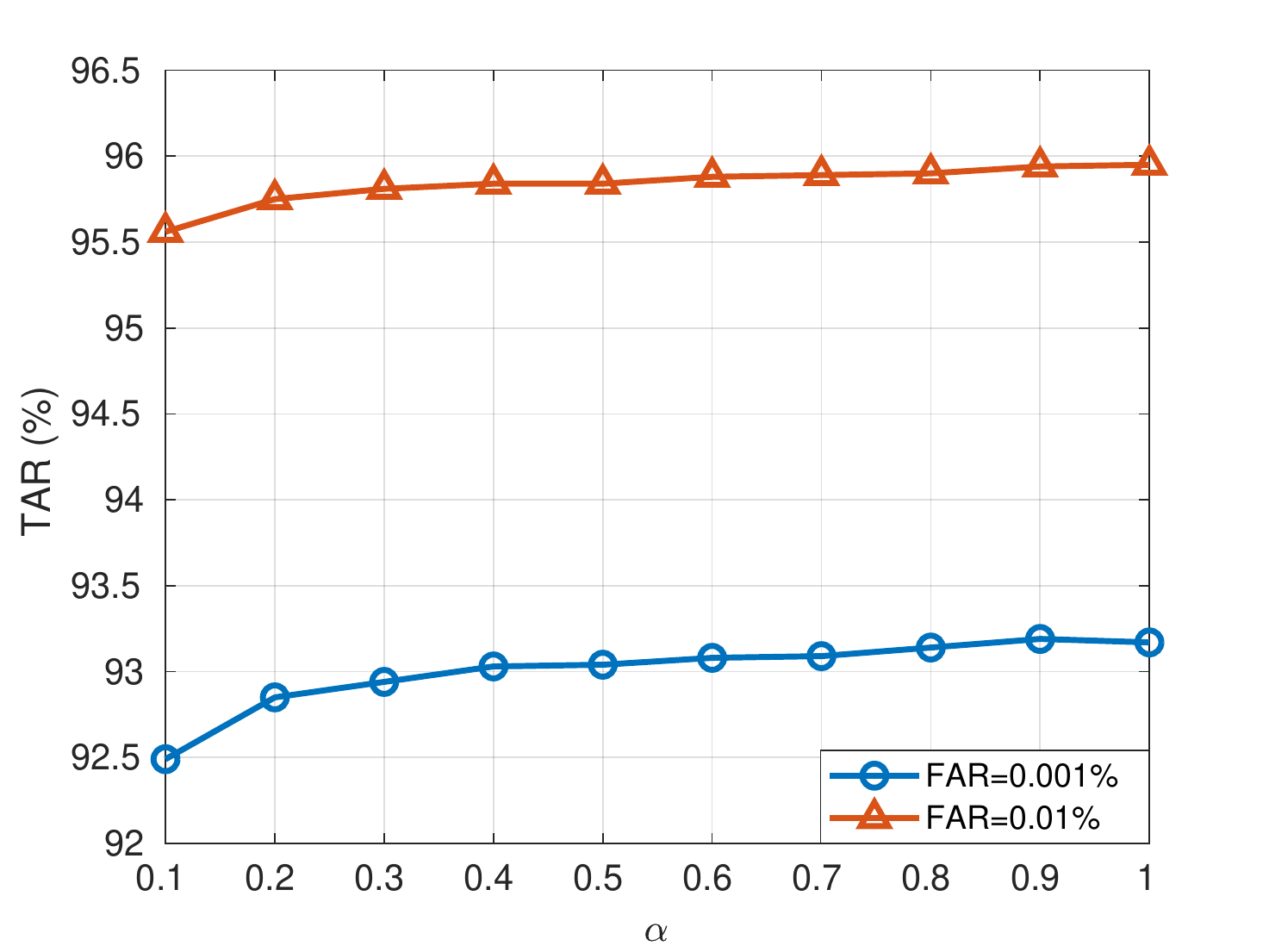}
\caption{The mean performance of different values of hyper-parameter $\alpha$ on five folds on the Private ID-selfie dataset.}
\label{fig:alpha}
\end{figure}

Figure~\ref{fig:alpha} shows how average generalization performance changes along with $\alpha$. Here, we build the mini-batches with random ID-selfie pairs from different classes. Then, $\bwjb$ is chosen as the average feature of the ID and selfie sample. From the figure, it is clear that a larger $\alpha$ always leads to better performance and the accuracy peaks when $\alpha=1$, where we directly replace the weights with $\bwjb$ as in Equation~(\ref{eq:update_w_hard}). This is not surprising because most classes only have two samples, and thus there is actually no need to update the weights softly since $\alpha=1$ always leads to the most accurate estimation of the class distribution. A smaller $\alpha$ might be preferred in the case of a deeper dataset.

The results of three different choices of $\bwjb$ are shown in Table~\ref{tab:wbatch}. Using either ID features or selfie features alone leads to a lower performance compared to the averaged feature. This is different from the results of Zhu et al.~\cite{zhu2018large}, who found that initializing the classifier weights as ID features leads to the best performance. Such differences may come from different strategies of updating classifier weights since we are updating them dynamically instead of keeping them fixed from the start of training. As most classes have only two images, one from each domain, updating the weights using only one of them causes a biased loss on these images and hence discourages the network from learning better representations for one domain.

\begin{table}[t]
\footnotesize
\begin{center}
\caption{The mean (and s.d. of) performance for different choices of $\bwjb$ based on 5-fold cross-validation on the Private ID-selfie dataset.}
\label{tab:wbatch}
\begin{tabularx}{\linewidth}{Xccc}
\toprule
$\bwjb$      & \multicolumn{3}{c}{True Accept Rate (\%)} \\ \cmidrule(lr){2-4}
                                     & FAR=$0.001\%$     & FAR=$0.01\%$  & FAR=$0.1\%$  \\
\midrule
ID features      & $87.35\pm 1.13$   & $92.87\pm 0.97$       & $96.17\pm 0.51$ \\
Selfie features  & $88.01\pm 1.51$   & $92.96\pm 1.02$       & $96.04\pm 0.68$       \\
Average         & $\mathbf{93.16\pm 0.85}$ & $\mathbf{95.95\pm 0.54}$ & $\mathbf{97.51\pm 0.40}$ \\
\bottomrule
\end{tabularx}
\end{center}
\end{table}

\subsection{Data Sampling}
\label{sec:exp:sampling}

Since our data can be categorized in two ways: identity and source, it raises the question of how we should sample the images for mini-batches during training. Here, we compare three different sampling methods: (1) image-wise random sampling, (2) random pairs from different classes, and (3) random ID-selfie pairs from different classes (proposed). Method (1) is commonly used for training classification loss functions, while method (2) is commonly used by metric learning methods because they require genuine pairs within the mini-batch for training. For (1) and (2), the classifier weights are updated with the average feature of the samples in each class. The corresponding results are shown in Table~\ref{tab:wbatch}. As one can see, random-image sampling works slightly better than random-pair sampling, which is consistent with the results on general face recognition. This is because random-pair leads to different sampling chances for images in different classes, and the model will be biased towards samples in small classes, which are sampled more frequently. However, in spite of this problem, random ID-selfie pairs still work slightly better than random-image, which shows that a balanced parameter learning of the two domains is important in our problem. These results imply that one should investigate how to further improve the performance by simultaneously solving the class imbalance problem (or long-tail problem) and domain imbalance problem.

\begin{table}[t]
\footnotesize
\begin{center}
\caption{The mean (and s.d. of) performance of different sampling methods for building mini-batches based on 5-fold cross-validation on the Private ID-selfie dataset.}
\label{tab:sampling}
\begin{tabularx}{\linewidth}{Xccc}
\toprule
Sampling                  & \multicolumn{3}{c}{True Accept Rate (\%)} \\ \cmidrule(lr){2-4}
                                     & FAR=$0.001\%$     & FAR=$0.01\%$  & FAR=$0.1\%$  \\
\midrule
Random images             & $92.77\pm 1.05$   & $95.71\pm 0.67$       & $97.41\pm 0.48$ \\
Random pairs              & $92.66\pm 0.93$   & $95.63\pm 0.70$       & $97.27\pm 0.43$       \\
Random ID-selfie pairs    & $\mathbf{93.16\pm 0.85}$ & $\mathbf{95.95\pm 0.54}$ & $\mathbf{97.51\pm 0.40}$ \\
\bottomrule
\end{tabularx}
\end{center}
\end{table}

\subsection{Parameter Sharing}
\label{sec:exp:param_sharing}
To evaluate the effect of shared parameters vs. domain-specific parameters, we constrain a subset of the parameters in the sibling networks to be shared between ID and selfie domains and compare the performances. Here, we consider both the case of shared low-level parameters and the case of shared high-level parameters. In particular, we compare the cases where modules ``Conv1'', ``Conv1-3'' and ``Conv1-5'' in the network are shared for learning low-level parameters. Then, we repeat the experiments by sharing the modules ``FC" and ``Conv5 + FC" and ``Conv4-5 + FC" for high-level parameters. Here ``Conv i-j" means the $i^{th}$ to $j^{th}$ convolutional modules and ``FC" means the fully connected layer for feature extraction. The results are shown in Table~\ref{tab:domain_features}.

\begin{table}[t]
\footnotesize
\begin{center}
\caption{The mean (and s.d. of) performance of constraining different modules of the sibling networks to be shared. ``All'' indicates a single model for both domains while ``None'' means that the parameters for the two domains are completely independent.}
\label{tab:domain_features}
\begin{tabularx}{\linewidth}{Xccc}
\toprule
Shared Modules  & \multicolumn{3}{c}{True Accept Rate (\%)} \\ \cmidrule(lr){2-4}
                                     & FAR=$0.001\%$     & FAR=$0.01\%$  & FAR=$0.1\%$  \\

\midrule
None            & $93.07\pm 0.91$   & $95.86\pm 0.56$       & $97.45\pm 0.39$       \\\hline
Conv 1          & $93.08\pm 0.95$   & $95.86\pm 0.58$       & $97.47\pm 0.40$       \\
Conv 1-3        & $93.13\pm 0.88$   & $95.84\pm 0.57$       & $97.46\pm 0.42$       \\
Conv 1-4        & $93.11\pm 0.85$   & $95.85\pm 0.57$       & $97.43\pm 0.41$       \\\hline
FC              & $\mathbf{93.16\pm 0.85}$ & $95.95\pm 0.54$ & $\mathbf{97.51\pm 0.40}$ \\
Conv 5 + FC     & $93.14\pm 0.89$   & $95.96\pm 0.55$       & $97.48\pm 0.43$       \\
Conv 4-5 + FC   & $93.10\pm 0.85$   & $\mathbf{95.97\pm 0.57}$       & $97.46\pm 0.42$ \\\hline
All             & $92.91\pm 0.92$   & $95.81\pm 0.63$       & $97.40\pm 0.43$       \\
\bottomrule
\end{tabularx}
\end{center}
\end{table}

From Table~\ref{tab:domain_features} we can see that the performance does not vary a lot with the parameter sharing. This is partially because our base model learns highly transferable features, whose parameters already provides a good initialization for all these modules. In particular, sharing low-level parameters does not have a clear impact on the performance, but constraining a shared bottleneck does lead to a slight improvement in both accuracy and standard deviation at all false accept rates. Then, the performance decreases when we constrain more high-level parameters to be shared. Furthermore, sharing all parameters leads to worse performance when compared with others. We can conclude, indeed, that there exists a small domain difference between ID photos and selfie photos and learning domain-specific parameters is helpful for recognition of photos across the two domains.

\subsection{Comparison with Static Weight Imprinting}
\label{sec:exp:static}
In~\cite{zhu2018large}, Zhu et al. used ID features as fixed classifier weights to run fine-grained training and showed performance improvement from the original representation. We regard such a method as static weight imprinting. The advantage of static weight imprinting is that one can extract the features simultaneously from all classes to update the classifier weights. However, they not only result in extra computational cost but also fail to capture the global distribution during the training. We compare our dynamic weight imprinting method with static imprinting methods. In particular, we consider two cases of static imprinting: (1) updating weights only at the beginning of fine-tuning and keeping them fixed during training and (2) updating weights every two epochs. For static methods, we extract the features of a random ID-selfie pair from every class and use their average vector for updating the weights. The results are shown in Table~\ref{tab:static}. It can be noted to see that periodical updating outperforms fixed weights, since fixed weights fail to keep up with the feature distribution. Better performance should be expected if we update the static weights more frequently. However, it is important to note that periodical updating also introduces more computational cost as we need to extract tens of thousands of features every time we update the weights. In comparison, DWI has almost zero extra computational cost, yet it leads to even better performance than periodical updating, indicating that the weights under the proposed DWI are able to keep up-to-date and capture the global distribution accurately.

\begin{table}[t]
\footnotesize
\begin{center}
\caption{The mean (and s.d. of) performance of static weight imprinting and dynamic weight imprinting based on 5-fold cross-validation on the Private ID-selfie dataset. ``Static-fixed'' updates all the weights at the beginning of fine-tuning. ``Static-periodical'' updates all the weights every two epochs. For the proposed ``DWI'', all the weights are randomly initialized.}
\label{tab:static}
\begin{tabularx}{\linewidth}{Xccc}
\toprule
Weight Update             & \multicolumn{3}{c}{True Accept Rate (\%)} \\ \cmidrule(lr){2-4}
                                     & FAR=$0.001\%$     & FAR=$0.01\%$  & FAR=$0.1\%$  \\
\midrule
Static - fixed            & $90.97\pm 1.01$   & $94.76\pm 0.64$     & $96.91\pm 0.46$   \\
Static - periodical       & $92.95\pm 0.85$   & $95.88\pm 0.50$     & $97.43\pm 0.39$   \\
\textit{DWI}                   & $\mathbf{93.16\pm 0.85}$ & $\mathbf{95.95\pm 0.54}$ & $\mathbf{97.51\pm 0.40}$ \\
\bottomrule
\end{tabularx}
\end{center}
\end{table}

\subsection{Comparison with Different Loss Functions}
\label{sec:exp:losses}

In this section, we evaluate the effect of different loss functions for fine-tuning on the Private ID-selfie dataset. We do not delve into the choice of loss function for training the base model since it is not relevant to the main topic of this paper. We compare the proposed DIAM-Softmax with three classification-based embedding learning loss functions: Softmax, A-Softmax~\cite{liu2017sphereface} and AM-Softmax~\cite{wang2018additive} and three other metric learning loss functions: contrastive loss~\cite{chopra2005learning}, triplet loss~\cite{schroff2015facenet} and the MPS loss proposed in our prior work~\cite{shi2018docface}. The classification-based loss functions are able to capture global information and thus have achieved state-of-the-art performances on general face recognition problems~\cite{wang2018additive} while the metric learning loss functions are shown to be effective on very large datasets~\cite{schroff2015facenet}. To ensure a fair comparison, we implement all the loss functions in Tensorflow and keep the experimental settings same except that AM-Softmax is trained for twice as long. The results are shown in Table~\ref{tab:losses}.

\begin{figure}[t]
\centering
\includegraphics[width=\linewidth]{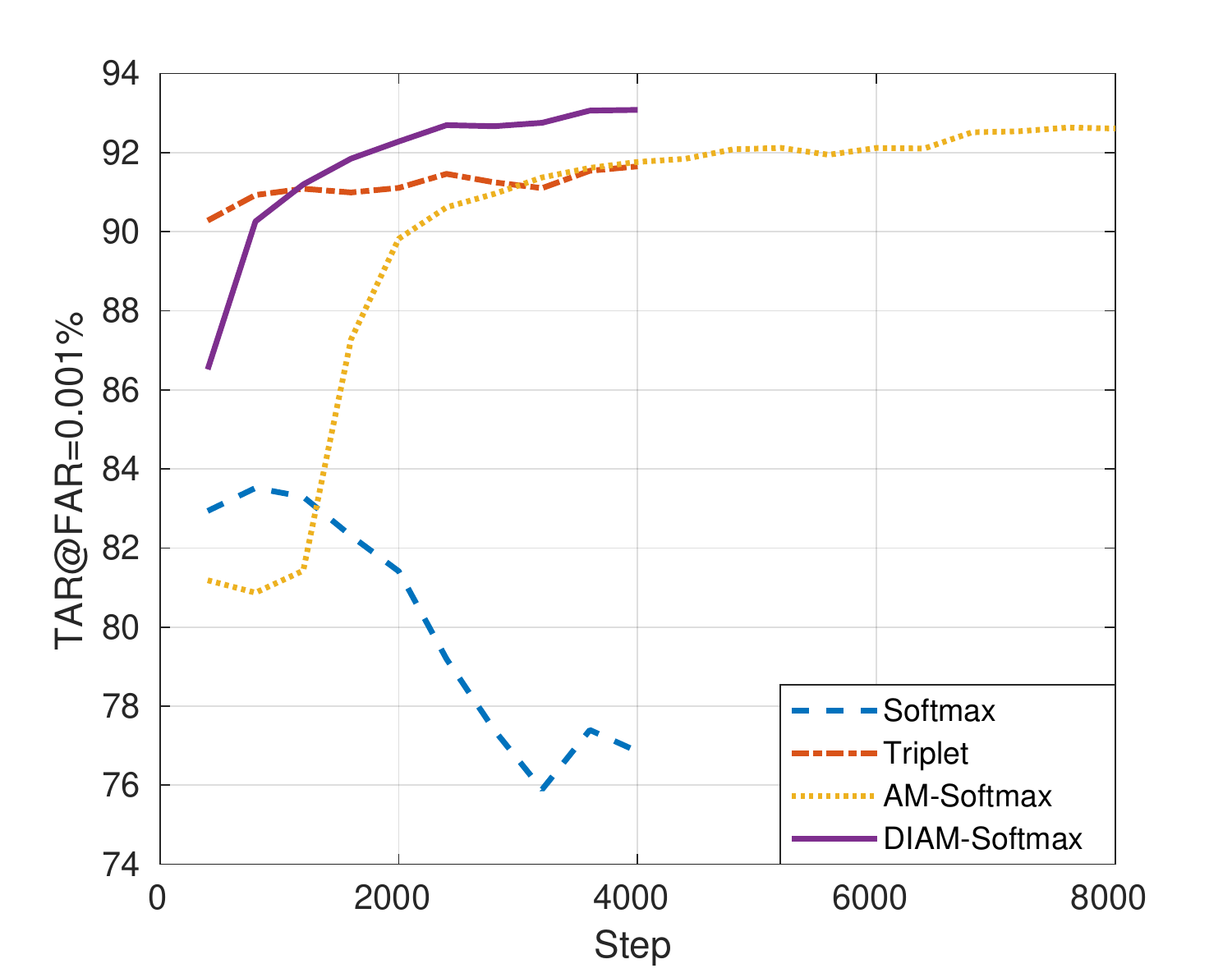}
\caption{Generalization performance of different loss functions during training. The x-axis is the number of training steps and the y-axis is the average TAR@FAR=$0.001\%$ of five folds at the corresponding step.}
\label{fig:tar_losses}
\end{figure}

\begin{table}[t]
\footnotesize
\begin{center}
\caption{The mean (and s.d. of) performance of different loss functions based on 5-fold fold cross-validation on the Private ID-selfie dataset. ``N/C'' means not converged. The proposed method is shown in italic style.}
\label{tab:losses}
\begin{tabularx}{\linewidth}{Xccc}
\toprule
Method   & \multicolumn{3}{c}{True Accept Rate (\%)} \\ \cmidrule(lr){2-4}
                                     & FAR=$0.001\%$     & FAR=$0.01\%$  & FAR=$0.1\%$  \\

\midrule
Base Model                              & $77.69\pm 2.02$   & $85.95\pm 1.67$       & $92.43\pm 0.98$   \\
Softmax                                 & $83.51\pm 1.76$   & $90.14\pm 1.44$       & $94.53\pm 0.81$   \\
A-Softmax~\cite{liu2017sphereface}      & N/C     & N/C       & N/C   \\
AM-Softmax~\cite{wang2018additive}      & $92.53\pm 1.09$   & $95.57\pm 0.57$       & $97.23\pm 0.42$   \\
Contrastive~\cite{chopra2005learning}   & $91.13\pm 1.65$   & $95.05\pm 0.77$       & $97.18\pm 0.47$   \\
Triplet~\cite{schroff2015facenet}       & $91.68\pm 1.21$   & $95.42\pm 0.70$       & $97.26\pm 0.45$   \\
MPS~\cite{chopra2005learning}           & $91.79\pm 1.16$   & $95.43\pm 0.65$       & $97.27\pm 0.44$   \\
\hline
\textit{DIAM-Softmax}   & $\mathbf{93.16\pm 0.85}$ & $\mathbf{95.95\pm 0.54}$ & $\mathbf{97.51\pm 0.40}$ \\
\bottomrule
\end{tabularx}
\end{center}
\end{table}

From Table~\ref{tab:losses}, one can see that our base model already achieves quite high performance (TAR of $92.43\%\pm 0.98$ at FAR of $0.1\%$) on the target dataset without any fine-tuning, indicating that the features learned on general face datasets are highly transferable, but it is significantly lower than its performance on general face datasets such as LFW due to the discrepancy between the characteristics of face images in these two tasks. Clear improvement can be observed after fine-tuning with most of the loss functions. To gain more insight, we plot the TAR-step curves (the x-axis is the number of training steps and the y-axis is the mean TAR@FAR=$0.001\%$ of five folds at that step) in Figure~\ref{fig:tar_losses}. We only pick out the representative loss functions for clarity of the plot. From Figure~\ref{fig:tar_losses}, we can see that Softmax overfits heavily just after two epochs\footnote{Each epoch is about $300$ steps}, while the metric learning loss is more stable and quick to converge. Although AM-Softmax performs better than metric learning methods, it converges so slowly that we have to train it for twice as many steps. Notice that this result is not contradictory with our prior work~\cite{shi2018docface}, where we found AM-Softmax perform poorly on fine-tuning, because we only allowed an equally limited number of training steps in~\cite{shi2018docface}. In comparison with the slow convergence of AM-Softmax, with the proposed weight updating method, the ``DIAM-Softmax'' not only converges faster, it is also robust against overfitting and generalizes much better than all competitors.

\subsection{Comparison with existing methods}
\label{sec:exp:private}

\begin{figure}[t]
\centering
\subfigure[False accept pairs]{
    \includegraphics[width=\linewidth]{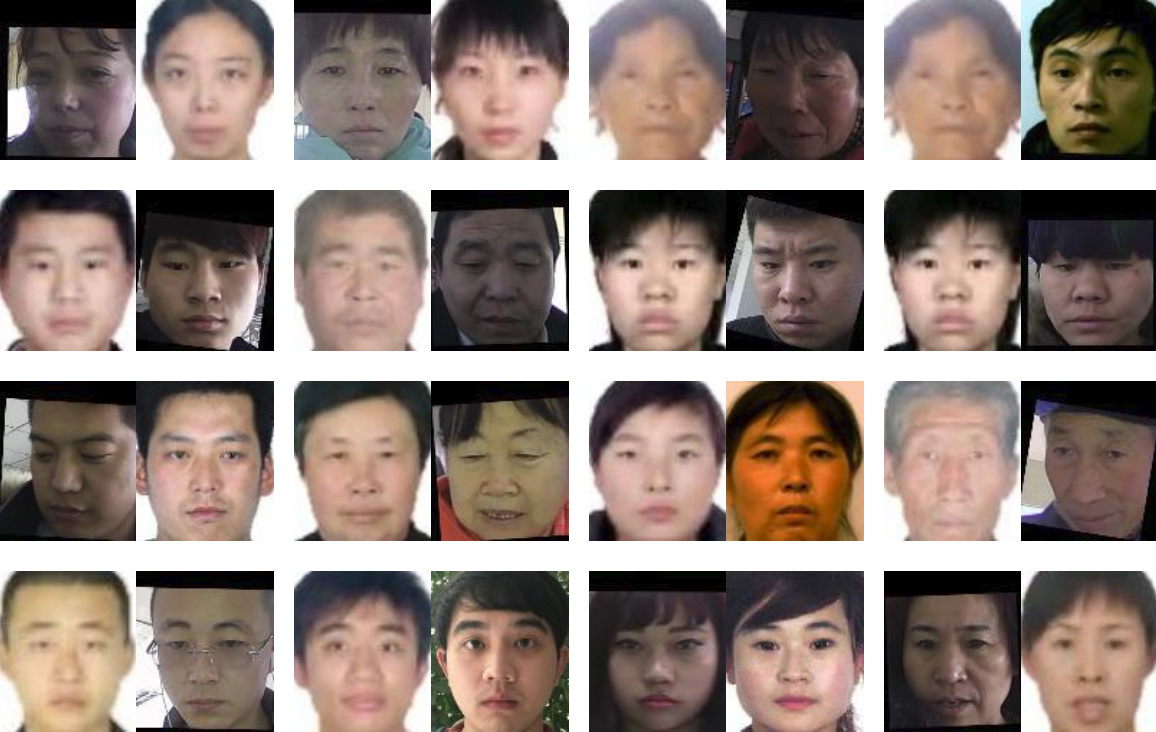}
}
\subfigure[False reject pairs]{
    \includegraphics[width=\linewidth]{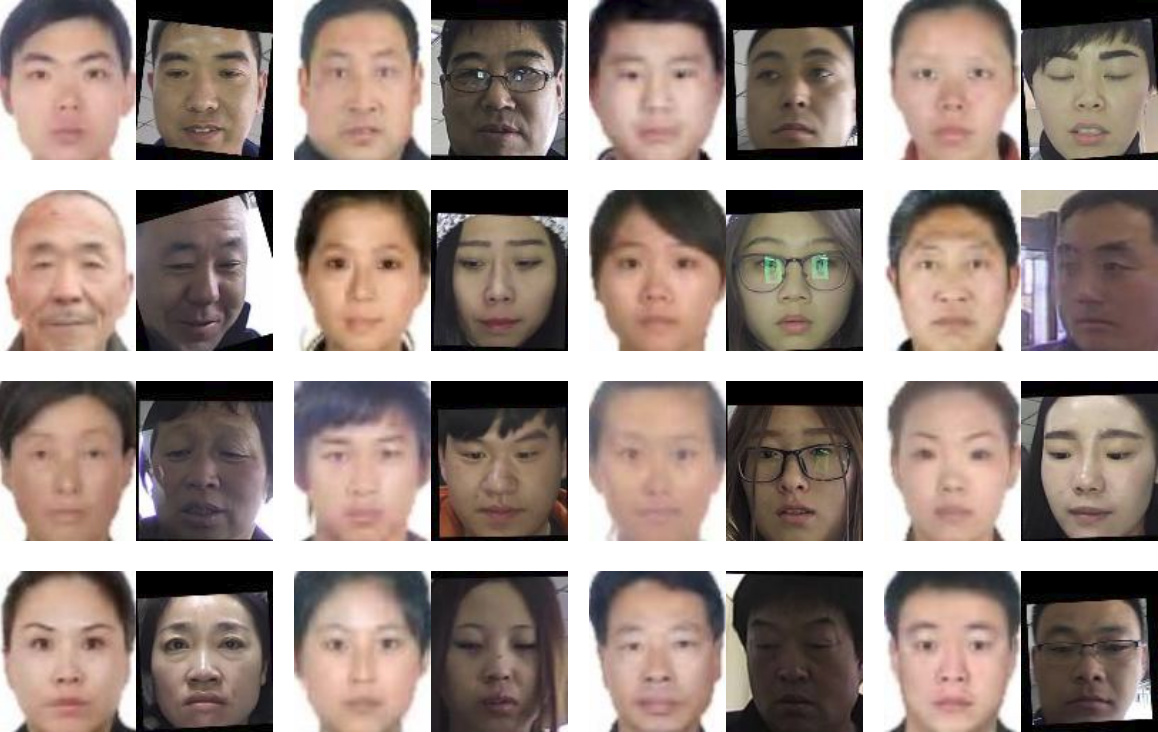}
}
\caption{Examples of falsely classified images by our model on the Private ID-selfie dataset at FAR = $0.001\%$.}
\label{fig:sample_failure_private}
\end{figure}

In this subsection, we evaluate the performance of other face matchers on the Private ID-selfie dataset to compare with our method. To the best of our knowledge, there are no existing public face matchers in the domain of ID-selfie matching. Although Zhu et al. developed a system on 2.5M ID-selfie training pairs, both their system and training data are not in the public domain and therefore we cannot compare their system with the proposed method on our dataset. Therefore, we compare our approach with state-of-the-art general face matchers to evaluate the efficacy of our system on the problem of ID-selfie matching. To make sure our experiments are comprehensive enough, we compare our method not only with two Commercial-Off-The-Shelf (COTS) face matchers, but also two state-of-the-art open-source CNN face matchers, namely CenterFace\footnote{\url{https://github.com/ydwen/caffe-face}}~\cite{wen2016discriminative} and SphereFace\footnote{\url{https://github.com/wy1iu/sphereface}}~\cite{liu2017sphereface}. During the five-fold cross-validation, because these general face matchers cannot be retrained, only the test split is used. The results are shown in Table~\ref{tab:cross_val}. Performances of the two open-source CNN matchers, CenterFace and SphereFace, are below par on this dataset, much worse than their results on general face datasets~\cite{wen2016discriminative}\cite{liu2017sphereface}. Although our base model performs better, it still suffers from a large drop in performance compared to its performance on general face datasets. It can be concluded that general CNN face matchers cannot be directly applied to the ID-selfie problem because the characteristics of ID-selfie images are different than those of general face datasets and a domain-specific modeling is imperative. A commercial state-of-the-art face recognition system, COTS-2, performs closer to our fine-tuned model. However, since the face dataset used to train COTS-2 is proprietary, it is difficult to conclude whether a general commercial face matcher can work well on this problem. In fact, from Table~\ref{tab:cross_val}, another commercial face matcher, COTS-1, performs much worse on this dataset.

\begin{table}[t]
\footnotesize
\begin{center}
\caption{The mean (and s.d. of) performance of different matchers on the private ID-selfie dataset. The ``base model" is only trained on MS-Celeb-1M. The model \textit{DocFace+} has been fine-tuned on training splits. Our models are shown in italic style. The two COTS and two public CNN face matchers are not retrained on our dataset.}
\label{tab:cross_val}
\begin{tabularx}{\linewidth}{Xccc}
\toprule
Method   & \multicolumn{3}{c}{True Accept Rate (\%)} \\ \cmidrule(lr){2-4}
                                     & FAR=$0.001\%$     & FAR=$0.01\%$  & FAR=$0.1\%$  \\

\midrule
COTS-1                                       & $58.62\pm 2.30$          & $68.03\pm 2.32$       & $78.48\pm 1.99$   \\
COTS-2                                       & $91.53\pm 1.96$          & $94.41\pm 1.84$       & $96.50\pm 1.78$ \\
CenterFace~\cite{wen2016discriminative}      & $27.37\pm 1.27$          & $41.38\pm 1.43$       & $59.29\pm 1.42$       \\
SphereFace~\cite{liu2017sphereface}          & $7.96\pm 0.68$   & $21.15\pm 1.63$   & $50.76\pm 1.55$       \\ \hline
\textit{Base model}                         & $77.69\pm 2.02$   & $85.95\pm 1.67$   & $92.43\pm 0.98$   \\
\textit{DocFace+}    & $\mathbf{93.16\pm 0.85}$ & $\mathbf{95.95\pm 0.54}$ & $\mathbf{97.51\pm 0.40}$ \\
\bottomrule
\end{tabularx}
\end{center}
\end{table}

\subsection{Evaluation on Public-IvS}
\label{sec:exp:public}

In~\cite{zhu2018large}, Zhu et al. released a \textit{simulated} ID-selfie dataset for open evaluation. The details and example images of this dataset are given in Section~\ref{sec:dataset:public}. Here, we test our system as well as the previous public matchers on this dataset for comparison. Among all the photos, we were able to successfully align $5,500$ images with MTCNN. Assuming that the subjects are cooperative and no failure-to-enroll would happen in real applications, we only test on the aligned images. The DocFace+ model is trained on the entire Private ID-selfie dataset. Hence, no cross-validation is needed here. The results are shown in Table~\ref{tab:test_public}.
\begin{table}[t]
\footnotesize
\begin{center}
\caption{Evaluation results on Public-IvS dataset. 
The model \textit{DocFace+} has been fine-tuned on the entire Private ID-selfie dataset and the performance of~\cite{zhu2018large} is reported in their paper. Our model is shown in italic style.}
\label{tab:test_public}
\begin{tabularx}{\linewidth}{Xccc}
\toprule
Method   & \multicolumn{3}{c}{True Accept Rate (\%)} \\ \cmidrule(lr){2-4}
                                     & FAR=$0.001\%$     & FAR=$0.01\%$  & FAR=$0.1\%$  \\

\midrule
COTS-1                                       & $83.78$                      & $89.92$                 & $92.90$   \\
COTS-2                                       & $\mathbf{94.74}$                & $97.03$           & $97.88$   \\
CenterFace~\cite{wen2016discriminative}      & $35.97$                & $53.30$           & $69.18$       \\
SphereFace~\cite{liu2017sphereface}          & $53.21$                & $69.25$           & $83.11$       \\
Zhu et al.~\cite{zhu2018large}               & $93.62$                & $\mathbf{97.21}$           & $\mathbf{98.83}$      \\ \hline
\textit{DocFace+}                             & $91.88$                & $96.48$           & $98.40$      \\

\bottomrule
\end{tabularx}
\end{center}
\end{table}
From the table, it can be seen that all the competitors perform better on this dataset than on our private dataset, indicating that this is an easier task. Since this dataset is a simulated ID-selfie dataset with web photos (see Section~\ref{sec:dataset:public}), a plausible explanation for these results is that there is a slight covariate shift between this dataset and our dataset. Therefore, our DocFace+ model, which is fine-tuned on the Private ID-selfie dataset, performs lower on this dataset. Notice that although Zhu et al.~\cite{zhu2018large} performs higher than our model, they used a training dataset of $2.5$M ID-selfie pairs while our training dataset (Private ID-selfie) only contains about $120$K images. Overall, we can still conclude that our model generalizes well on this dataset and outperforms most competitors significantly.

\begin{figure}[t]
\centering
\subfigure[False accept pairs]{
    \includegraphics[width=\linewidth]{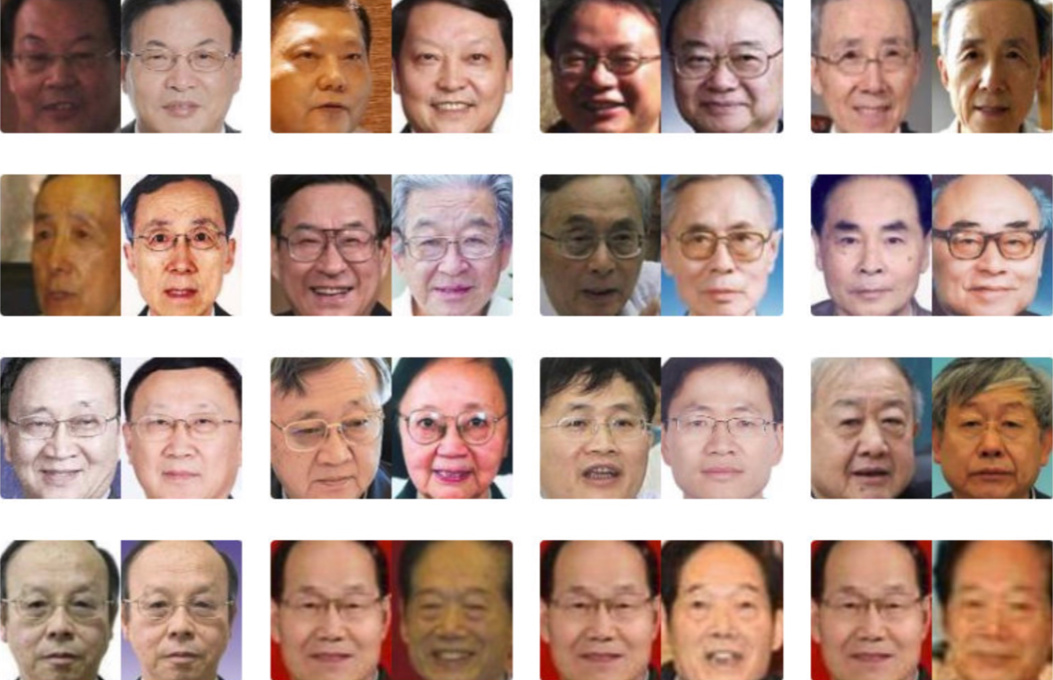}
}
\subfigure[False reject pairs]{
    \includegraphics[width=\linewidth]{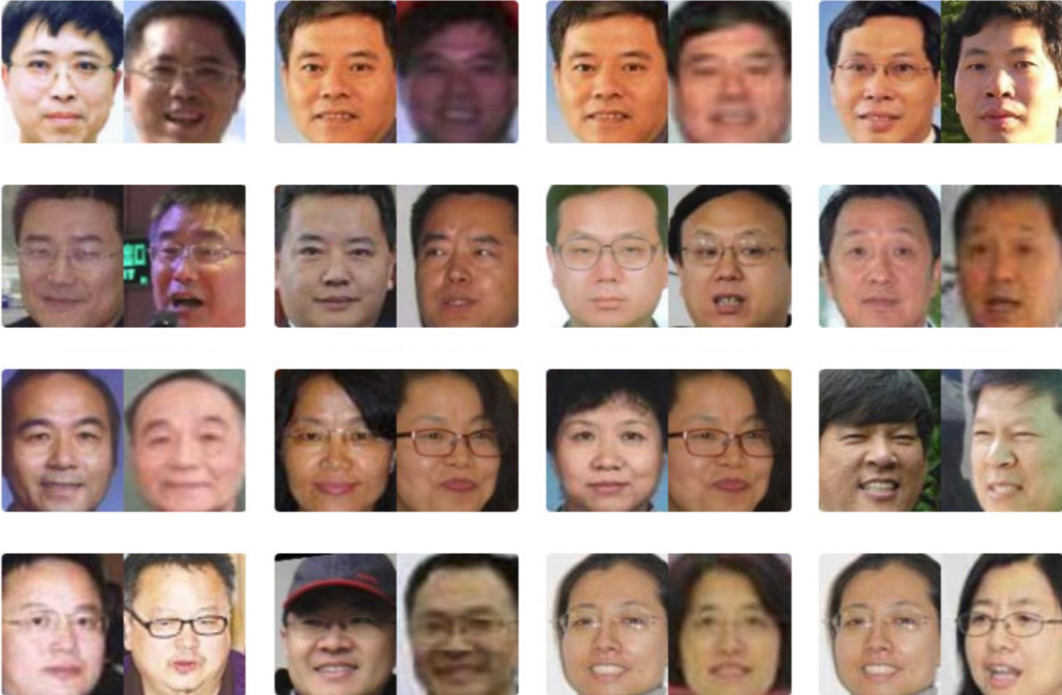}
}
\caption{Examples of falsely classified images by our model on the Public ID-selfie dataset at FAR = $0.001\%$.}
\label{fig:sample_failure_public}
\end{figure}

\section{Conclusions}
In this paper, we propose a new system, named DocFace+, for matching ID document photos to selfies. The transfer learning technique is used where a base model for unconstrained face recognition is fine-tuned on a private ID-selfie dataset. A pair of sibling networks with shared high-level modules are used to model domain-specific parameters. Based on our observation of the weight-shift problem of classification-based embedding learning loss functions on shallow datasets, we propose an alternative optimization method, called dynamic weight imprinting (DWI) and a variant of AM-Softmax, DIAM-Softmax. Experiments show that the proposed method not only helps the loss converge much faster but also leads to better generalization performance. A comparison with static weight imprinting methods confirms that DWI is capable of capturing the global distribution of embeddings accurately. Different sampling methods are studied for mini-batch construction and we find that a balanced sampling between the two domains is most helpful for learning generalizable features. We compare the proposed system with existing general face recognition systems on our private dataset and see a significant improvement with our system, indicating the necessity of domain-specific modeling of ID-selfie data. Finally, we compare the performance of different matchers on the Public-IvS dataset and find that although there is a covariate shift, our system still generalizes well.

\section*{Acknowledgement} 
This work is partially supported by Visa, Inc. We also thank ZKTeco for providing the ID-Selfie datasets.

{
\small
\bibliographystyle{IEEEtran}
\bibliography{ref}
}

\newcommand{\biospace}{\vspace{-0.5in}}
\biospace
\begin{IEEEbiography}[{\includegraphics[width=1in,height=1.25in,clip,keepaspectratio]{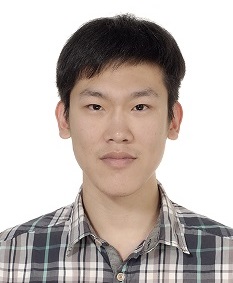}}]{Yichun~Shi} received his B.S degree in the Department of Computer Science and Engineering at Shanghai Jiao Tong University in 2016. He is now working towards the Ph.D. degree in the Department of Computer Science and Engineering at Michigan State University. His research interests include pattern recognition and computer vision.
\end{IEEEbiography}

\biospace
\begin{IEEEbiography}[{\includegraphics[width=1in,height=1.25in,clip,keepaspectratio]{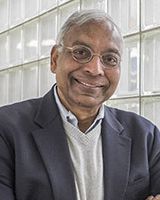}}]{Anil~K.~Jain} is a University distinguished professor in the Department of Computer Science and Engineering at Michigan State University. His research interests include pattern recognition and biometric authentication. He served as the editor-in-chief of the IEEE Transactions on Pattern Analysis and Machine Intelligence (1991-1994) and as a member of the United States Defense Science Board. He was elected to the National Academy of Engineering and Indian National Academy of Engineering.
\end{IEEEbiography}

\end{document}